\documentclass[letterpaper, 10 pt, conference]{ieeeconf}  

\IEEEoverridecommandlockouts                              

\usepackage{graphics} 
\usepackage{epsfig} 
\usepackage{mathptmx} 
\usepackage{times} 
\usepackage{amsmath} 
\usepackage{amssymb}  
\usepackage{algorithm, algpseudocode}
\usepackage{xcolor}
\usepackage{subcaption}
\usepackage{booktabs}
\usepackage{multirow}
\usepackage{cite}
\usepackage{siunitx}
\usepackage{booktabs}
\usepackage{graphicx}
\usepackage{multirow}
\usepackage{threeparttable}
\usepackage{url}
\usepackage{hyperref}

\title{\LARGE \bf
MR-MPCC: Model Predictive Contouring Control for Robotic Manipulator Tracking under Moving Reference Frames 
}
\date{today}

\author{
Chonghao Cheng\textsuperscript{1},
Tiancheng Li\textsuperscript{1},
Jonathan Woolfrey\textsuperscript{1},
Liang Zhao\textsuperscript{2},
and Shoudong Huang\textsuperscript{1}%
\thanks{\textsuperscript{1}Robotics Institute, University of Technology Sydney, Sydney, NSW, Australia. Email: Chonghao.Cheng@student.uts.edu.au}%
\thanks{\textsuperscript{2}School of Informatics, University of Edinburgh, Edinburgh, UK.}%
\thanks{This work is supported by National Health and Medical Research Council (NHMRC) Ideas Grant (ID:2029811) and the Royal Society of Edinburgh.}
}

\begin{document}

\maketitle
\thispagestyle{empty}
\pagestyle{empty}

\begin{abstract}
Maintaining small path-tracking errors under moving reference frames is essential for robotic manipulation tasks with tight execution tolerances, such as machining and surgical automation. Conventional time-indexed controllers, including tracking Model Predictive Control (MPC), rely on a fixed temporal correspondence with the reference trajectory and may therefore accumulate path-relative error under abrupt reference-frame motion or timing mismatch. This paper presents MR-MPCC, a Model Predictive Contouring Control (MPCC) framework for velocity-controlled robotic manipulators with 6-DoF end-effector pose regulation under Moving-Reference disturbances. A hierarchical control architecture is adopted, in which an upper-level MPCC optimizes path progress while penalizing contouring and orientation errors, and a constrained resolved motion rate control (RMRC) layer maps the commanded Cartesian twist to feasible joint velocities under joint-level and safety constraints. MR-MPCC is implemented in a ROS2--MuJoCo pipeline and validated on a UR5e platform under moving-reference disturbances, where the path frame is perturbed and tracked by electromagnetic sensors. Compared with PI control and time-indexed tracking MPC under matched disturbance, MR-MPCC achieves lower geometric path-following error and smaller deviations in the moving path frame, demonstrating improved path-following performance under reference-frame motion.
\end{abstract}

\section{Introduction}

Accurate end-effector motion control is fundamental to robotic
manipulation in industrial assembly, surgical automation, and machining~\cite{siciliano2009handbook,ji2019industrial}.
In such tasks, performance is usually determined by geometric fidelity to a prescribed path rather than by strict adherence to a preassigned timing law~\cite{kapitanyuk2017guiding}. The control objective is therefore to minimize spatial deviation
from the path in the presence of disturbances and modeling uncertainty.

Conventional controllers such as proportional--integral--derivative (PID) control provide intuitive and computationally efficient solutions~\cite{GARPINGER2014568}, but do not explicitly predict future system evolution or systematically handle state and input constraints.
Model Predictive Control (MPC) addresses these issues by optimizing
future control actions over a finite horizon~\cite{camachoMPC2007,hewingLearningBasedMPC2020}.
However, standard MPC formulations are time-indexed and rely on a fixed
time parameterization of the reference~\cite{faulwasser2009nonlinear}. Under disturbances that induce
timing mismatch or abrupt motion of the reference frame, time-indexed
tracking can accumulate geometric error, leading to degraded robustness,
especially when simultaneous position and orientation regulation is
required.

Model Predictive Contouring Control (MPCC) offers an alternative by reformulating tracking relative to the reference path~\cite{lamModelPredictiveContouring2010}. It decomposes the tracking deviation into a contouring component orthogonal to the path and a lag component along the path, allowing geometric path-following accuracy and path progress to be regulated separately.
Despite these advantages, its application to robotic manipulators remains in its infancy, with only a small number of recent works beginning to explore this domain~\cite{Yoon2025ReactiveMP}.

\begin{figure}
    \centering
    \includegraphics[width=0.9\linewidth]{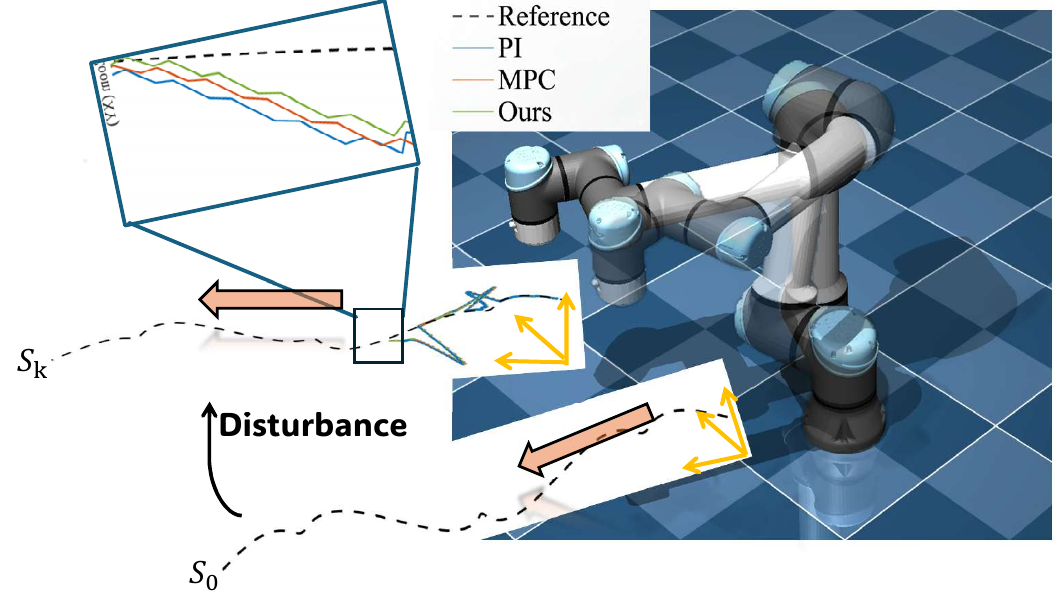}
    \caption{MuJoCo simulation snapshot illustrating moving-reference disturbances applied to the path frame. An augmented figure shows a local disturbance segment, revealing reduced overshoot and faster return to the reference trajectory under our framework compared with PI and MPC.}
    \label{fig:placeholder}
\end{figure}

Extending MPCC to robotic manipulators is challenging because nonlinear kinematics, and joint-level constraints complicate real-time task-space control~\cite{craigIntroductionRoboticsMechanics2018}. The challenge increases when the reference frame moves abruptly, since stable joint-level control does not necessarily ensure small path-relative errors. In applications such as surgical milling and high-accuracy assembly, these errors can affect task quality and safety. This motivates direct regulation of path-relative error under reference motion and joint constraints.

This paper presents MR-MPCC, a task-space MPCC framework for velocity-controlled manipulators with full 6-DoF end-effector pose regulation. The method combines MPCC with constrained RMRC to generate feasible joint-velocity commands in real time. It is implemented in a ROS2--MuJoCo pipeline (Fig.~\ref{fig:placeholder}) and validated on a physical UR5e, where reference-frame motion is measured using electromagnetic tracking and the controller operates at 500 Hz. Under matched paths and disturbance, MR-MPCC achieves lower contouring error than PI control and time-indexed tracking MPC in both simulation and physical experiments.

The main contributions are:
(i) a task-space MPCC formulation for full 6-DoF manipulator path following under moving reference frames;
(ii) a unified simulation and physical evaluation pipeline for reference-frame disturbances; and
(iii) a comparison with PI control and time-indexed tracking MPC of contouring, lag, and rotational tracking errors.

\section{Related Work}

\subsection{Manipulator Control and Disturbance Rejection}

Robotic manipulators are subject to modeling uncertainties and external disturbances that can degrade
tracking accuracy and closed-loop stability. Classical controllers such as PID and feedback linearization
provide reliable baseline performance, but disturbance attenuation often depends on careful tuning and
model fidelity~\cite{GARPINGER2014568}.

Robust control techniques explicitly shape uncertainty margins to improve disturbance rejection, including
sliding-mode control~\cite{pham2025robustadaptivefuzzysliding} and $H_\infty$ designs~\cite{FIGUEREDO2021109817}.
In practice, these methods can trade tracking smoothness or responsiveness for robustness, and they do not
directly address the timing sensitivity induced by a time-parameterized reference.

Iterative learning control leverages task repetitiveness to reduce tracking errors across
repeated executions, which has been widely adopted in machining and contouring applications~\cite{bristow2006survey,liFiveaxisContourError2023,moore1992iterative}.
However, its effectiveness diminishes when disturbances are non-repetitive or when the task
conditions change rapidly between trials.


Overall, these approaches strengthen robustness via feedback compensation or repetition-based
error refinement, but they do not explicitly optimize future motion under constraints.
This motivates predictive control formulations that can incorporate constraints
and anticipate disturbance effects.

\subsection{MPC for Manipulators}


MPC is attractive for manipulator systems since finite-horizon optimization can explicitly encode multivariable dynamics, actuation limits, and safety constraints, enabling motion objectives to be pursued subject to requirements such as joint-limit enforcement, collision avoidance, and pose-constrained path or tunnel following~\cite{woolfreyPredictiveEndEffectorControl2021,camachoMPC2007,astudillo2022tunnelfollowing}.


Robust MPC variants, including tube-based designs, further account for model uncertainty and external
disturbances. For instance, Yao \textit{et al}.\ \cite{yao2023disturbance_tmmpc} combined disturbance observers with tube MPC to improve
tracking robustness under uncertainty and input constraints, while Kang
\textit{et al}.\ \cite{kang2023sm_tube_mpc} developed a sliding-mode adaptive tube MPC to enhance robustness and constraint satisfaction
under dynamic uncertainties.
Learning-augmented MPC~\cite{hewingLearningBasedMPC2020} has also been studied to improve model fidelity in unstructured settings.

A recurring limitation of time-indexed MPC is sensitivity to timing mismatch and reference-motion effects:
execution delays or moving-reference disturbances can preserve time alignment while accumulating spatial
deviation from the underlying geometric path. Spatially aware formulations such as MPCC mitigate this issue
by directly penalizing contouring and lag errors relative to the path parameterization, decoupling path fidelity
from time parameterization.

\subsection{Evolution of MPCC}

MPCC~\cite{lamModelPredictiveContouring2010} was introduced to balance geometric path accuracy
against traversal progress in contouring systems by formulating tracking in path coordinates
and explicitly penalizing contouring and lag errors.
Subsequent studies~\cite{9873215} extended the formulation to multi-axis motion platforms to achieve
high-precision contour tracking under actuation constraints.

MPCC later gained traction in robotics for high-performance trajectory tracking~\cite{linigerOptimizationbasedAutonomousRacing2015a}, where optimizing
the path progress online enables adaptive speed regulation while directly minimizing geometric
deviation from the reference path.
To support general spatial motion, recent works~\cite{romeroModelPredictiveContouring2022,romero2022time} formalized consistent contouring and lag error
definitions in three dimensions, enabling MPCC on arbitrary 3D paths.
Constraint-aware variants have further incorporated safety structures into the optimization,
such as corridor-constrained MPCC~\cite{jiCMPCCCorridorbasedModel2021} for aggressive flight along spatial paths.

The application of MPCC to robotic manipulators remains relatively limited. Yoon \textit{et al}.\ \cite{Yoon2025ReactiveMP} proposed a reactive MPCC formulation augmented with control barrier functions for obstacle avoidance, self-collision prevention, and singularity handling, and demonstrated real-time performance on a 7-DoF Panda manipulator. Their work primarily addresses safe and feasible path following in constrained environments. In contrast, this paper focuses on accurate manipulator tracking when external disturbances alter the reference frame, requiring online compensation for relative robot--reference motion.

\section{Preliminaries}
This section introduces the path parameterization and the time-indexed tracking formulation used as a baseline reference. It then defines the contouring and lag errors and derives the MPCC problem that jointly optimizes the control input and the path progress.






\subsection{Reference Path}

The reference path is defined by a sequence of Cartesian waypoints
$\{\mathbf{p}^d_k\}_{k=0}^{M}$ and interpolated into a smooth curve
$\mathbf{p}^d(s)\in\mathbb{R}^3$, $s\in[0,1]$.
A predefined timing law specifies the time-parameterized reference
\begin{align}
s_k = s(t_k),\  \dot{s}=v_{\mathrm{ref}} , \ \mathbf{p}^d_k := \mathbf{p}^d(s_k).
\end{align}

\subsection{Time-Parameterized Tracking MPC}




At each sampling instant, MPC computes a control sequence by solving
the finite-horizon optimization problem
\begin{equation}
\label{eq:mpc_standard}
\begin{aligned}
    \min_{\{\mathbf{u}_{k+i}\}_{i=0}^{N-1}}
    \quad
    \sum_{i=0}^{N-1}
    \Big(
        &\|
        \mathbf{p}_{k+i}
        -
        \mathbf{p}^d(s_{k+i})
        \|_{\mathbf{Q}}^2
        +
        \|
        \mathbf{u}_{k+i}
        -
        \mathbf{u}^d(s_{k+i})
        \|_{\mathbf{R}}^2
    \Big)
    \\
    \text{s.t.}\quad
    &\mathbf{p}_{k+i+1}
    =
    f(\mathbf{p}_{k+i},\mathbf{u}_{k+i}),
    \\
    &\mathbf{p}_{k+i} \in \mathcal{P}, \quad
    \mathbf{u}_{k+i} \in \mathcal{U}.
\end{aligned}
\end{equation}
where $f(\cdot,\cdot)$ denotes a general nonlinear discrete-time model and $\mathbf{Q}, \mathbf{R}$ denote the weighted matrices.


\subsection{Model Predictive Contouring Control}

MPCC removes the fixed time-parameterization and instead regulates
geometric deviation from the path while optimizing the progression
along it.

\begin{figure}[t]
  \centering
  \includegraphics[width=1.0\linewidth]{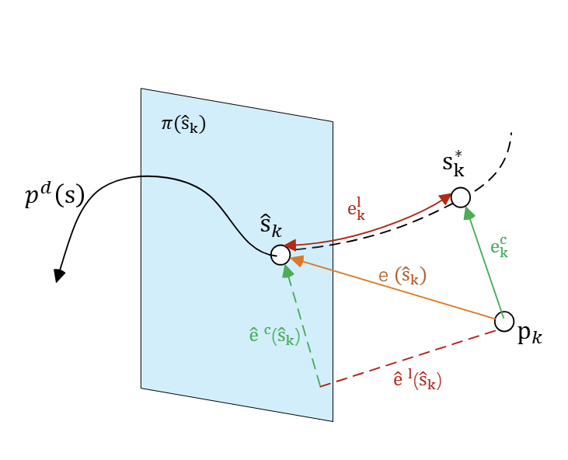}
  \caption{Contouring/lag error and error estimation in MPCC. 
  The reference path $\mathbf{p}^d(s)$ (black) is parameterized by arc length. 
  The closest-point parameter $s_k^\ast$ defines the true errors with respect to the system position $\mathbf{p}_k$: 
  the contouring error $\mathbf{e}^c_k$ (green) and the lag error $\mathbf{e}^l_k$ (red). 
  To avoid solving the closest-point projection at each step, MPCC propagates an estimated progress variable $\hat{s}_k$, 
  yielding approximate errors $\hat{\mathbf{e}}^c(\hat{s}_k)$ and $\hat{\mathbf{e}}^l(\hat{s}_k)$ computed at $\mathbf{p}^d(\hat{s}_k)$. 
  The shaded plane $\pi(\hat{s}_k)$ indicates the local linearization/projection neighborhood.}
  \label{fig:estimation}
\end{figure}

\textbf{Closest-point definition.} For a given system position $\mathbf{p}_k$, the closest point on the
reference path is defined as
\begin{equation}
    s_k^\ast
    =
    \arg\min_s
    \|\mathbf{p}_k - \mathbf{p}^d(s)\|.
\end{equation}
The associated position error is
\begin{equation}
    \mathbf{e}(s_k^\ast)
    =
    \mathbf{p}_k
    -
    \mathbf{p}^d(s_k^\ast).
\end{equation}
The error can be decomposed into components with respect to the
\emph{local} path tangent at the closest point.
In particular, the scalar lag error is the projection of the position
error onto the unit tangent $\mathbf{t}(s_k^\ast)$,
\begin{equation}
    e_k^l
    =
    \mathbf{t}(s_k^\ast)^\top
    \mathbf{e}(s_k^\ast),
\end{equation}
while the contouring error is the residual in the orthogonal complement
of $\mathbf{t}(s_k^\ast)$,
\begin{equation}
    \mathbf{e}_k^c
    =
    \mathbf{e}(s_k^\ast)
    -
    e_k^l \mathbf{t}(s_k^\ast).
\end{equation}
As illustrated in Fig.~\ref{fig:estimation}, the lag error corresponds to the tangential component along the path, while the contouring error represents the orthogonal deviation.




\textbf{MPCC optimization problem.} The finite-horizon MPCC problem is
\begin{equation}
\begin{aligned}
    \min_{\{\mathbf{u}_{k+i+1},v_{s,k+i+1}\}}
    \sum_{i=0}^{N-1}
    \Big(
        \|\mathbf{e}_{k+i}^c\|_{\mathbf{Q}_c}^2
        &+
        (e_{k+i}^\ell)^2{Q_\ell}
        \\+
        \|\mathbf{u}_{k+i}-&\mathbf{u}^d(s_{k+i})\|_{\mathbf{R}}^2
        -
        q_s v_{s,k+i}
    \Big)
    \\
    \text{s.t.}
    \quad
    \mathbf{p}_{k+i+1}
    &=
    f(\mathbf{p}_{k+i},\mathbf{u}_{k+i}),
    \\
    \hat{s}_{k+i+1}
    &=
    \hat{s}_{k+i}
    +
    v_{s,k+i}\Delta t,
    \\
    \mathbf{p}_{k+i}\in&\mathcal{P},
    \quad
    \mathbf{u}_{k+i}\in\mathcal{U}.
\end{aligned}
\end{equation}
where $\mathbf{Q}_c \in \mathbb{R}^{3\times3},~Q_\ell \in \mathbb{R}$ denote the contouring and lag error weights, respectively.
MPCC builds upon the tracking MPC formulation by augmenting the optimization problem with the path progress dynamics, replacing the predefined timing law with an optimized virtual velocity.

\section{Hierarchical MPCC-based Framework for Manipulators}

The framework we proposed follows a two-layer hierarchy.
At the upper layer, a task-space MPCC controller generates a desired
end-effector twist together with a path progress command based on the
current robot state.
At the lower layer, this task-space command is realized through joint
velocities by solving a constrained quadratic program that enforces
kinematic feasibility and safety constraints.




\subsection{Problem Formulation}

Let the end-effector pose be given by position ${}^{W}\mathbf{p}(t)\in\mathbb{R}^3$ and unit quaternion
${}^{W}\mathbf{q}(t)\in\mathbb{S}^3$.
A reference path is specified in the path-attached frame $\mathcal{F}_D$ as
\begin{equation}
{}^{D}\mathbf{r}(s)=
\begin{bmatrix}
{}^{D}\mathbf{p}_r(s)\\
{}^{D}\mathbf{q}_r(s)
\end{bmatrix},
\ \ s\in[0,1],
\end{equation}
where ${}^{D}\mathbf{p}_r(s)\in\mathbb{R}^3$ and ${}^{D}\mathbf{q}_r(s)\in\mathbb{S}^3$.

Moving-reference disturbances are injected through the time-varying pose of $\mathcal{F}_D$ with respect to
$\mathcal{F}_W$, triggered by the progress variable $s(t)$:
\begin{align}
{}^{W}\mathbf{p}_{D}(t)&= {}^{W}{\mathbf{p}}_{D}(t-1) + \mathbf{d}^{p}(s(t)),\nonumber\\
{}^{W}\mathbf{q}_{D}(t)&= \mathbf{d}^{\omega}(s(t))\otimes {}^{W}{\mathbf{q}}_{D}(t-1),
\end{align}
where $\mathbf{d}^{p}(s) \in \mathbb{R}^3$ and $\mathbf{d}^{\omega}(s) \in \mathbb{S}^3$ are disturbances introduced at progress location $s$.

The induced moving reference in $\mathcal{F}_W$ is
\begin{align}
{}^{W}\mathbf{p}^d(t,s)&= \mathbf{R}\!\left({}^{W}\mathbf{q}_{D}(t)\right)\,{}^{D}\mathbf{p}_r(s) + {}^{W}\mathbf{p}_{D}(t),\nonumber \\
{}^{W}\mathbf{q}^d(t,s)&= {}^{W}\mathbf{q}_{D}(t)\otimes {}^{D}\mathbf{q}_r(s),
\end{align}
where $\mathbf{R}(\mathbf{q})$ maps a unit quaternion to a rotation matrix.


The control objective is to use the commanded end-effector twist
${}^{W}\mathbf{u}(t)=\big[({}^{W}\mathbf{v}(t))^\top,({}^{W}\boldsymbol{\omega}(t))^\top\big]^\top$ to regulate $({}^{W}\mathbf{p}(t),{}^{W}\mathbf{q}(t))$ toward
$({}^{W}\mathbf{p}^d(t,s),{}^{W}\mathbf{q}^d(t,s))$ while selecting the progression variable $s(t)$ and
satisfying actuator and safety constraints.







\subsection{Upper Layer: Task-Space MPCC Formulation}

\textbf{Frame convention.} Predicted states and control inputs are expressed in the world frame $\mathcal{F}_W$.
Geometric errors are evaluated in the path-attached frame $\mathcal{F}_D$ using the current estimate of the
path-frame pose $\big({}^{W}\mathbf{p}_{D,k},{}^{W}\mathbf{q}_{D,k}\big)$, which is held constant over the prediction horizon. At each control step $k$, the predicted state is
$[({}^{W}\mathbf{p}_k)^\top,({}^{W}\mathbf{q}_k)^\top,s_k]^\top$,
where ${}^{W}\mathbf{p}_k\in\mathbb{R}^3$, ${}^{W}\mathbf{q}_k\in\mathbb{S}^3$, and $s_k\in[0,1]$.

\textbf{Prediction model.} The kinematics in $\mathcal{F}_W$ are
\begin{align}
{}^{W}\mathbf{p}_{k+1} &= {}^{W}\mathbf{p}_k + {}^{W}\mathbf{v}_k \Delta t, \nonumber\\
{}^{W}\mathbf{q}_{k+1} &= \exp\!\left(\tfrac{1}{2}{}^{W}\boldsymbol{\omega}_k \Delta t\right)\otimes {}^{W}\mathbf{q}_k,\nonumber\\
s_{k+1} &= s_k + v_{s,k} \Delta t,
\end{align}
where the input twist is ${}^{W}\mathbf{u}_k=[({}^{W}\mathbf{v}_k)^\top,({}^{W}\boldsymbol{\omega}_k)^\top]^\top\in\mathbb{R}^6$.

\textbf{Reference twist.} Let ${}^{W}\mathbf{u}^d(s)=[({}^{W}\mathbf{v}^d(s))^\top,({}^{W}\boldsymbol{\omega}^d(s))^\top]^\top$
denote the feedforward twist of the reference pose path expressed in $\mathcal{F}_W$, and define
${}^{W}\mathbf{u}^d_{k+i}={}^{W}\mathbf{u}^d(s_{k+i})$.

\textbf{World-to-path mapping.} At step $k$, the estimated path-frame pose $({}^{W}\mathbf{p}_{D,k},{}^{W}\mathbf{q}_{D,k})$ defines the rigid transform from $\mathcal{F}_W$ to $\mathcal{F}_D$, which is held constant over the prediction horizon. Predicted poses and twists, together with the feedforward twist ${}^{W}\mathbf{u}^d(s)$, are re-expressed in $\mathcal{F}_D$ via $\mathbf{R}({}^{W}\mathbf{q}_{D,k})^\top$ to compute errors with respect to ${}^{D}\mathbf{r}(s)$.

\textbf{Geometric position errors.} The reference path is defined in $\mathcal{F}_D$ as
${}^{D}\mathbf{p}_r(s)$ with unit tangent
${}^{D}\mathbf{t}(s)=\frac{{}^{D}\mathbf{p}_r'(s)}{\|{}^{D}\mathbf{p}_r'(s)\|}$.
Then
\begin{align}
{}^{D}\mathbf{e}_{k+i} &= {}^{D}\mathbf{p}_{k+i} - {}^{D}\mathbf{p}_r(s_{k+i}),\nonumber\\
e_{k+i}^\ell &= \left({}^{D}\mathbf{t}(s_{k+i})\right)^\top {}^{D}\mathbf{e}_{k+i},\nonumber\\
{}^{D}\mathbf{e}_{k+i}^c &=
\left(\mathbf{I}-{}^{D}\mathbf{t}(s_{k+i})\,{}^{D}\mathbf{t}(s_{k+i})^\top\right){}^{D}\mathbf{e}_{k+i}.
\end{align}

\textbf{Orientation error.} Let ${}^{D}\mathbf{q}_r(s)$ denote the reference orientation in $\mathcal{F}_D$. The multiplicative error is
\begin{align}
&\mathbf{q}_{e,k+i} = \left({}^{D}\mathbf{q}_r(s_{k+i})\right)^{-1}\otimes {}^{D}\mathbf{q}_{k+i},\nonumber\\
&\boldsymbol{\phi}_{k+i}=2\,\mathbf{q}_{e,k+i,\mathrm{vec}}.
\end{align}

\textbf{Optimal control problem.} Over a horizon of $N$ steps, MPCC solves
\begin{equation}
\label{eq:mpcc_task_ocp}
\begin{aligned}
\min_{\{{}^{W}\mathbf{u}_{k+i},\,v_{s,k+i}\}_{i=0}^{N-1}}
\quad
\sum_{i=0}^{N-1}
\Big(
\|&{}^{D}\mathbf{e}_{k+i}^c\|_{\mathbf{Q}_c}^2
+ (e_{k+i}^\ell)^2 Q_\ell
+ \|\boldsymbol{\phi}_{k+i}\|_{\mathbf{Q}_o}^2\\
+& \|{}^{D}\mathbf{u}_{k+i}-{}^{D}\mathbf{u}^d_{k+i}\|_{\mathbf{R}_u}^2
- q_s v_{s,k+i}
\Big) \\
\text{s.t.}\quad
&{}^{W}\mathbf{p}_{k+i+1} = {}^{W}\mathbf{p}_{k+i} + {}^{W}\mathbf{v}_{k+i}\Delta t, \\
&{}^{W}\mathbf{q}_{k+i+1} =
\exp\!\left(\tfrac{1}{2}{}^{W}\boldsymbol{\omega}_{k+i} \Delta t\right)\otimes {}^{W}\mathbf{q}_{k+i}, \\
&s_{k+i+1} = s_{k+i} + v_{s,k+i}\Delta t, \\
&{}^{W}\mathbf{u}_{\min} \le {}^{W}\mathbf{u}_{k+i} \le {}^{W}\mathbf{u}_{\max}, \\
&v_{s,\min} \le v_{s,k+i} \le v_{s,\max}, \\
&0 \le s_{k+i+1} \le 1,\qquad i=0,\dots,N-1.
\end{aligned}
\end{equation}
where $\mathbf{Q}_c$, $Q_\ell$, $\mathbf{Q}_o$, and $\mathbf{R}_u$ weight contouring, lag, orientation, and input deviation, respectively, $q_s$ weights path progress, and only the first control pair $({}^{W}\mathbf{u}_k,v_{s,k})$ is applied.
\subsection{Lower Layer: Joint-Space Realization with Constraints}
\label{subsec:joint_space_realization}

The upper layer task-space MPCC outputs an end-effector twist command
${}^W\mathbf{u}_k\in\mathbb{R}^6$ expressed in the robot base frame.
Let $\boldsymbol{\theta}_k\in\mathbb{R}^n$ denote the joint configuration.
The differential kinematics relates joint velocities
$\dot{\boldsymbol{\theta}}_k\in\mathbb{R}^n$ to the twist as
\begin{equation}
    \mathbf{u}_k=\mathbf{J}(\boldsymbol{\theta}_k)\dot{\boldsymbol{\theta}}_k,
\end{equation}
where $\mathbf{J}(\boldsymbol{\theta}_k)\in\mathbb{R}^{6\times n}$ is the robot Jacobian.

\textbf{Constrained resolved motion rate control(RMRC).}
We compute a feasible joint-velocity command using an constrained RMRC-based QP~\cite{RMRC} that tracks the task-space command $\mathbf{u}_k$ while enforcing actuation and safety constraints.
The least-squares solution is
\begin{equation}
\label{eq:ik_ls_nominal}
    \dot{\boldsymbol{\theta}}_k^{\mathrm{ls}}
    =
    \arg\min_{\dot{\boldsymbol{\theta}}}
    \frac{1}{2}\|
        \mathbf{J}(\boldsymbol{\theta}_k)\dot{\boldsymbol{\theta}}-\mathbf{u}_k
    \|^2,
\end{equation}
which can be ill-conditioned near singular configurations.
To improve robustness, we incorporate singularity avoidance at both the objective and
constraint levels.

\textbf{Singularity avoidance via manipulability.}
We use the manipulability index \cite{Yoshikawa}
\begin{equation}\label{eq:Manipulability_rewrite}
    \mu(\boldsymbol{\theta})
    =
    \sqrt{\det\!\left(
        \mathbf{J}(\boldsymbol{\theta})\mathbf{J}(\boldsymbol{\theta})^\top
    \right)}.
\end{equation}

\textit{(i) Damped least squares.}
We regularize the RMRC using a configuration-dependent damping term,
\begin{equation}
\label{eq:ik_dls}
    \dot{\boldsymbol{\theta}}_k^{\mathrm{dls}}
    =
    \arg\min_{\dot{\boldsymbol{\theta}}}
    \frac{1}{2}\|
        \mathbf{J}(\boldsymbol{\theta}_k)\dot{\boldsymbol{\theta}}-\mathbf{u}_k
    \|^2
    +
    \frac{\lambda(\boldsymbol{\theta}_k)^2}{2}\|\dot{\boldsymbol{\theta}}\|^2,
\end{equation}
where $\lambda(\boldsymbol{\theta}_k)\ge 0$ increases as $\mu(\boldsymbol{\theta}_k)$ approaches
a threshold $\mu_{\min}$.

\textit{(ii) Manipulability barrier constraint.}
To prevent further loss of manipulability near $\mu_{\min}$, we impose \cite{kurtz2021control}
\begin{equation}
\label{eq:manipulability_barrier_rewrite}
    \nabla_{\boldsymbol{\theta}}\mu(\boldsymbol{\theta}_k)^\top \dot{\boldsymbol{\theta}}
    \ge
    -\gamma\big(\mu(\boldsymbol{\theta}_k)-\mu_{\min}\big),
\end{equation}
with $\gamma>0$, which enforces non-decreasing manipulability at the boundary and relaxes
smoothly away from it.

\textbf{Kinematic and safety constraints.}
In addition to singularity avoidance, we restrict joint velocities to satisfy actuator
limits and safety constraints.

\textit{Joint-velocity bounds with position and acceleration consistency.}
Joint velocities are constrained by elementwise bounds
\begin{equation}
\label{eq:joint_rate_bounds_rewrite}
    \dot{\boldsymbol{\theta}}_k^{\min}
    \le
    \dot{\boldsymbol{\theta}}
    \le
    \dot{\boldsymbol{\theta}}_k^{\max},
\end{equation}
where the effective limits are computed as
\begin{align}
\dot{\boldsymbol{\theta}}_k^{\min}
&=
\max\!\left(
\dot{\boldsymbol{\theta}}_{\min},\;
\frac{\boldsymbol{\theta}_{\min}-\boldsymbol{\theta}_k}{\Delta t},\;
\dot{\boldsymbol{\theta}}_{k-1}-\ddot{\boldsymbol{\theta}}_{\max}\Delta t
\right),\\
\dot{\boldsymbol{\theta}}_k^{\max}
&=
\min\!\left(
\dot{\boldsymbol{\theta}}_{\max},\;
\frac{\boldsymbol{\theta}_{\max}-\boldsymbol{\theta}_k}{\Delta t},\;
\dot{\boldsymbol{\theta}}_{k-1}+\ddot{\boldsymbol{\theta}}_{\max}\Delta t
\right),
\end{align}
with all \emph{max/min} operations applied elementwise.

\textit{Self-collision avoidance.}
Let $\sigma_{ij}(\mathbf{\theta})$ denote the minimum distance between a link pair $(i,j)$.
Linearizing around $\mathbf{\theta}_k$ gives
\begin{equation}
\label{eq:collision_linearization_rewrite}
    \dot \sigma_{ij}
    \approx
    \nabla_{\mathbf{\theta}} \sigma_{ij}(\mathbf{\theta}_k)^\top \dot{\mathbf{\theta}}.
\end{equation}
For distances below a safety threshold $\sigma_{\min}$, we enforce the velocity-damper constraint
\begin{equation}
\label{eq:collision_barrier_rewrite}
    \nabla_{\mathbf{\theta}} \sigma_{ij}(\mathbf{\theta}_k)^\top \dot{\mathbf{\theta}}
    \ge
    -\eta\big(\sigma_{ij}(\mathbf{\theta}_k)-\sigma_{\min}\big),
\end{equation}
with $\eta>0$.
Multiple link-pair constraints can be included by stacking inequalities of the form
\eqref{eq:collision_barrier_rewrite}.

\textbf{Final constrained RMRC.}
Combining the damped least-squares objective, the manipulability barrier, and the kinematic
and safety constraints yields the constrained RMRC-based QP:
\begin{equation}
\label{eq:constrained_ik_qp_rewrite}
\begin{aligned}
    \dot{\mathbf{\theta}}_k^\ast
    =
    \arg\min_{\dot{\mathbf{\theta}}}
    \;&
    \frac{1}{2}
    \|\mathbf{J}(\mathbf{\theta}_k)\dot{\mathbf{\theta}}
    - \mathbf{u}_k\|^2
    +
    \frac{\lambda(\mathbf{\theta}_k)^2}{2}
    \|\dot{\mathbf{\theta}}\|^2
    \\
    \text{s.t.}\;&
    \dot{\mathbf{\theta}}_{\min}
    \le
    \dot{\mathbf{\theta}}
    \le
    \dot{\mathbf{\theta}}_{\max},
    \\
    &
    \nabla_{\mathbf{\theta}}\mu(\mathbf{\theta}_k)^\top
    \dot{\mathbf{\theta}}
    \ge
    -\gamma\big(\mu(\mathbf{\theta}_k)-\mu_{\min}\big),
    \\
    &
    \nabla_{\mathbf{\theta}} \sigma_{ij}(\mathbf{\theta}_k)^\top
    \dot{\mathbf{\theta}}
    \ge
    -\eta\big(\sigma_{ij}(\mathbf{\theta}_k)-\sigma_{\min}\big).
\end{aligned}
\end{equation}
This constrained RMRC layer maps the MPCC twist ${}^W\mathbf{u}_k$ to a feasible joint velocity $\dot{\mathbf{\theta}}_k^\ast$ with joint-level singularity and collision avoidance.

\section{Experiment}

This section presents the simulation and physical evaluation of the proposed MPCC-based framework. Simulation results provide a controlled comparison under multiple disturbance profiles, while physical experiments on a UR5e manipulator assess the feasibility under moving-reference disturbances.

\subsection{Simulation}
\subsubsection{Simulation Setup}

PI, trajectory tracking MPC, and the proposed framework are evaluated in MuJoCo using a 6-DoF UR5e model under joint-velocity control.
All controllers operate at the same update rate and share the same kinematic model and joint-level constraints. For MPC and MPCC, QPs are solved online using a real-time solver \cite{woolfrey_simple_qp_2023}.
Results are averaged over five matched trials per configuration, using identical disturbance realizations across controllers.
Control parameters are summarized in Table~\ref{tab:control_parameters}.

\begin{table}[t]
\centering
\caption{MPCC parameters used in both simulation and physical experiments.}
\label{tab:control_parameters}
\begin{tabular}{l c}
\toprule
Control frequency $f_c$ & 500 Hz \\
Prediction horizon $N$ & 10 \\
Contouring weight $\mathbf{Q}_c$ & 500$\cdot\mathbf{I}_3$\\
Lag weight $Q_\ell$ &  20\\
Orientation weight $\mathbf{Q}_o$ & 40$\cdot \mathbf{I}_3$  \\
Input weight $\mathbf{R}_u$ &  0.01$\cdot \mathbf{I}_7$\\
\bottomrule
\end{tabular}
\end{table}

\subsubsection{Disturbance Models}
\label{sec:dist_models}

Disturbances are applied to the pose of the path-attached frame $\mathcal{F}_D$ used to express the reference,
\begin{equation}
{}^{W}\hat{\mathbf{T}}_{D,k}
=
{}^{W}\mathbf{T}_{D,k}\exp\!\big(\boldsymbol{\xi}^\wedge_k\big),\ \
\boldsymbol{\xi}_k=
\begin{bmatrix}
\mathbf{d}^{p}_k\\
\mathbf{d}^{\omega}_k
\end{bmatrix},
\end{equation}
where $\mathbf{d}^{p}_k\in\mathbb{R}^3$ and $\mathbf{d}^{\omega}_k\in\mathbb{R}^3$ are translational and rotation-vector perturbations.
We consider translation only (TOnly), rotation only (ROnly), and combined translation and rotation (RT) cases, with the disturbance triggered at the same nominal progress
$s^\star$ to align the disturbed path location across controllers.

\begin{figure}[htbp]
    \centering
    \includegraphics[width=\linewidth]{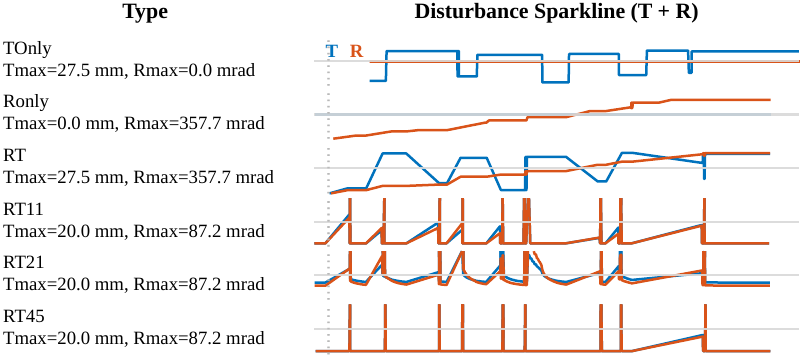}
\caption{Disturbance profiles used in the simulation. Each row corresponds to one disturbance configuration. Blue: translational component; orange: rotational component. The peak amplitudes for translation and rotation are on the left.}
    \label{fig:disturbance_visualization}
\end{figure}

For RT, let $\mathbf{d}_k=[(\mathbf{d}^{p}_k)^\top,(\mathbf{d}^{\omega}_k)^\top]^\top\in\mathbb{R}^6$ and generate temporally correlated transients via an impulse-driven ARMA$(p,q)$ time series model recursion,
\begin{equation}
\label{eq:arma_impulse_rt_piecewise}
\begin{aligned}
\mathbf{d}_k
&=
\sum_{i=1}^{p} \mathbf{A}_i \mathbf{d}_{k-i}
+
\sum_{j=0}^{q} \mathbf{B}_j \mathbf{s}_{k-j},\ \
\mathbf{s}_k=
\begin{cases}
\bar{\mathbf{d}}, & k=k_0,\\
\mathbf{0}, & \text{otherwise},
\end{cases}
\end{aligned}
\end{equation}
with $\bar{\mathbf{d}}=[(\bar{\mathbf{d}}^{p})^\top,(\bar{\mathbf{d}}^{\omega})^\top]^\top$.
Temporal structure is varied by the ARMA order $(p,q)$ and coefficients. Profiles are shown in Fig.~\ref{fig:disturbance_visualization}.

\subsubsection{Evaluation Metrics}
\label{sec:metrics}

All metrics are evaluated in the board-attached frame $\mathcal{F}_D$ and summarized in Table~\ref{tab:key_metrics}. We report contouring error as the primary measure of geometric path fidelity, along with lag and orientation errors; RMSE, MAE, and MAX are computed per trial and averaged over five matched trials.


\begin{table*}[t]
\centering
\caption{Tracking errors under path frame evaluation.}
\label{tab:key_metrics}
\scriptsize
\setlength{\tabcolsep}{4pt}
\renewcommand{\arraystretch}{0.8}

\begin{threeparttable}
\resizebox{\textwidth}{!}{%
\renewcommand{\arraystretch}{0.9}
\begin{tabular}{lll l
S[table-format=1.5, detect-all] S[table-format=1.5, detect-all] S[table-format=2.3, detect-all]
S[table-format=1.4, detect-all] S[table-format=2.3, detect-all]
S[table-format=2.3, detect-all] S[table-format=2.3, detect-all]}
\toprule
\multirow{2}{*}{Scenario} & \multirow{2}{*}{ARMA$(p,q)$} & \multirow{2}{*}{Trajectory} & \multirow{2}{*}{Method} &
\multicolumn{3}{c}{Contouring error [mm]} &
\multicolumn{2}{c}{Lag error [mm]} &
\multicolumn{2}{c}{Rotation error [mrad]} \\
\cmidrule(lr){5-7}\cmidrule(lr){8-9}\cmidrule(lr){10-11}
& & & & {RMSE} & {MAE} & {MAX} & {RMSE} & {MAX} & {RMSE} & {MAX} \\
\midrule

\multirow{18}{*}{RT} & \multirow{3}{*}{None} & \multirow{3}{*}{Spiral} & PI
& 5.3638 & 2.7572 & 52.613 & 4.5603 & 31.873 & 25.750 & 87.689 \\
& & & MPC
& 5.5058 & 2.6372 & 52.692 & 4.5335 & 32.038 & 24.988 & 87.724 \\
& & & Ours
& \bfseries 4.2794 & \bfseries 2.3627 & \bfseries 52.294 
& \bfseries 4.3819 & \bfseries 31.446
& \bfseries 24.778 & \bfseries 87.65 \\
\cmidrule(lr){2-11}

& \multirow{3}{*}{(1,1)} & \multirow{3}{*}{Spiral} & PI
& 5.5616 & 2.5324 & \bfseries 51.502 & 5.5164 & \bfseries 42.456 & 7.613 & 87.657 \\
& & & MPC
& 4.7092 & 2.2240 & 51.968 & \bfseries 5.4503 & 43.325 & 5.426 & \bfseries 87.655 \\
& & & Ours
& \bfseries 3.7197 & \bfseries 2.0321 & 51.784 & 5.4638 & 43.248 & \bfseries 2.856 & 87.689 \\
\cmidrule(lr){2-11}

& \multirow{9}{*}{(2,1)\textsuperscript{*}} & \multirow{3}{*}{Spiral} & PI
& 7.0666 & 3.4680 & \bfseries 51.809 & 5.5080 & 43.741 & 15.818 & \bfseries 87.798 \\
& & & MPC
& 6.0512 & 3.0546 & 51.829 & 5.4655 & 43.204 & 14.249 & 87.793 \\
& & & Ours
& \bfseries 5.5260 & \bfseries 3.0222 & 52.519 & \bfseries 5.4208 & \bfseries 43.058 & \bfseries 13.661 & 87.800 \\
\cmidrule(lr){3-11}

&  & \multirow{3}{*}{Flower} & PI
& 4.9478 & 1.9511 & 44.181 & \bfseries 0.0293 & \bfseries 7.4293 & 8.9011 & 3.6189 \\
& & & MPC
& 4.6197 & 1.9061 & 44.0510 & 1.1238 & 7.4413 & 8.8175 & 3.6229 \\
& & & Ours
& \bfseries 4.4782 & \bfseries 1.8576 & \bfseries 43.9480 & 1.0949 & 7.4390 & \bfseries 8.7314 & \bfseries 3.6176 \\
\cmidrule(lr){3-11}

& & \multirow{3}{*}{Helix} & PI
& 12.450  & 10.965  & 44.628
& 1.0107  & 22.462
& 2.8941  & 7.6417 \\
& & & MPC
& 12.373  & 10.914  & \bfseries 44.616
& 0.93783 & \bfseries 22.309
& \bfseries 2.8717 & 7.6959 \\
& & & Ours
& \bfseries 12.211 & \bfseries 10.587 & 44.626
& \bfseries 0.92331 & 22.487
& 2.8830  & \bfseries 7.2777 \\
\cmidrule(lr){2-11}

& \multirow{3}{*}{(4,5)} & \multirow{3}{*}{Spiral} & PI
& 5.0336 & 2.3623 & 51.842 & 5.3787 & \bfseries 40.980 & 6.729 & 87.699 \\
& & & MPC
& 4.6670 & 2.2343 & \bfseries 50.765 & \bfseries 5.3757 & 41.509 & \bfseries 5.786 & 87.681 \\
& & & Ours
& \bfseries 4.1935 & \bfseries 2.1512 & 52.040 & 5.6985 & 41.460 & 7.101 & \bfseries 87.675\\
\midrule

\multirow{3}{*}{ROnly} & \multirow{3}{*}{None} & \multirow{3}{*}{Spiral} & PI
& 4.0744 & 1.8340 & \bfseries 43.073 & \bfseries 3.0766& \bfseries 28.736 & 29.042 & \bfseries 87.678 \\
& & & MPC
& 5.7795 & 2.9085 & 49.006 & 5.4321 & 38.689 & \bfseries 26.679 & 87.713 \\
& & & Ours
& \bfseries 3.8892 & \bfseries 1.8161 & 43.107 & 3.0797 & 28.753 & 28.855 & 87.753 \\
\midrule

\multirow{3}{*}{TOnly} & \multirow{3}{*}{None} & \multirow{3}{*}{Spiral} & PI
& 1.8842 & 0.60846 & \bfseries 14.441 & \bfseries 0.9678 & \bfseries 17.684 & 0.701 & 1.5959 \\
& & & MPC
& 0.96572 & 0.40526 & 14.444 & 1.1939 & 18.927 & \bfseries 0.697 & \bfseries 1.4868 \\
& & & Ours
& \bfseries 0.70162 & \bfseries 0.34429 & 14.443 & 1.2726 & 17.786 & 0.707 & 1.5580 \\
\bottomrule

\end{tabular}%
}
\begin{tablenotes}[flushleft]
\scriptsize
\item[*] ARMA$(2,1)$ is selected as a representative correlated RT disturbance with moderate order and sufficient temporal structure. To isolate trajectory-dependent effects from changes in disturbance statistics, the multi-trajectory comparison is conducted under this fixed setting.
\end{tablenotes}
\end{threeparttable}
\end{table*}

\begin{figure}
    \centering
    \includegraphics[width=0.8\linewidth]{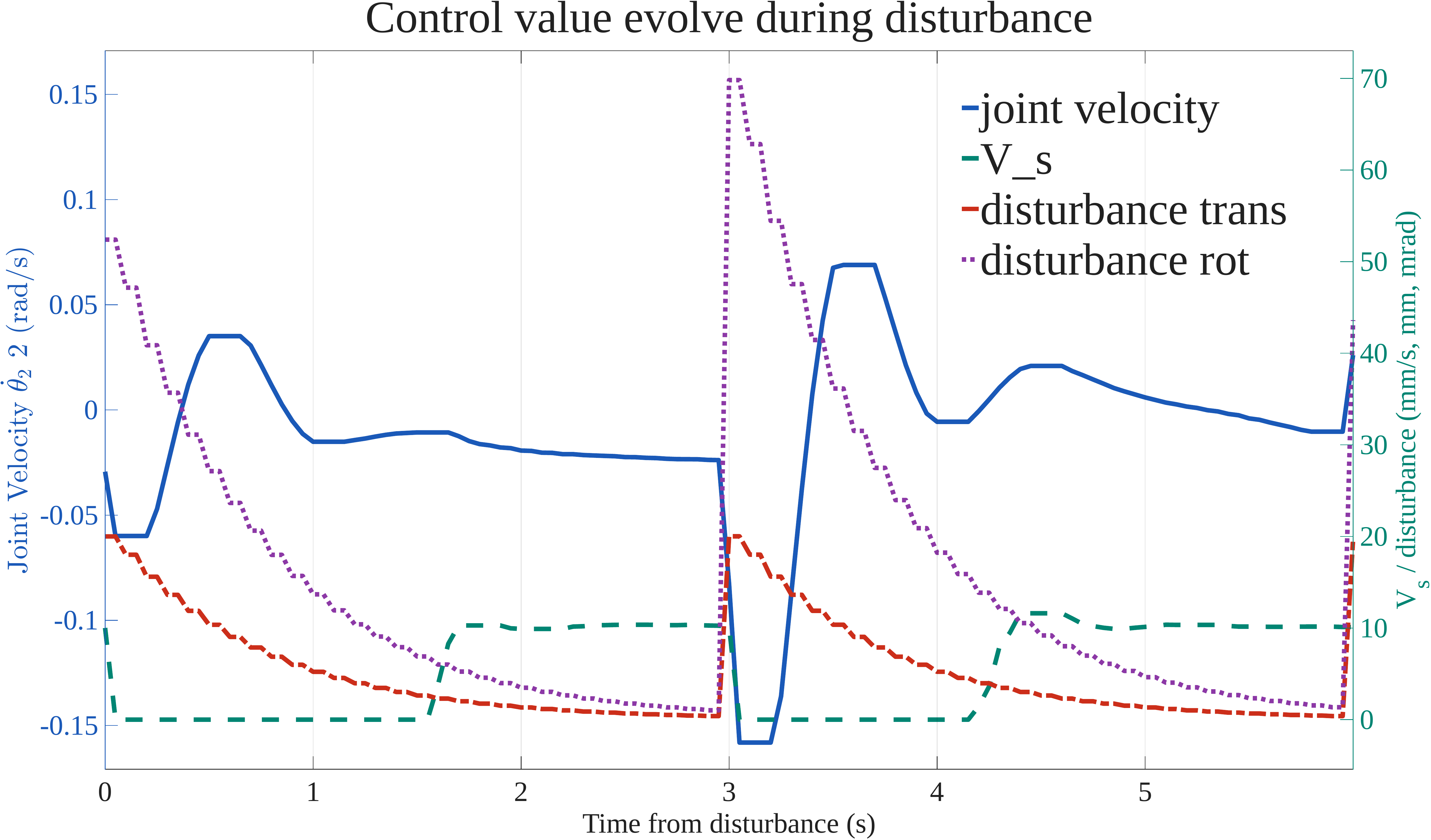}
    \caption{Joint-velocity response (blue solid) and our MPCC formulation progress rate $v_s$ (green dashed) after disturbance onset. MPCC suppresses the progress rate $v_s$ during large deviations, which reduces transient deviation in contouring error at the cost of increased lag error.}
    \label{fig:controlvalue}
\end{figure}

\subsubsection{Simulation Results}
\label{sec:sim_results}

Table~\ref{tab:key_metrics} reports PI, tracking MPC, and our MPCC-based framework under TOnly, ROnly, and RT disturbances. Multi-trajectory comparisons fix the correlated RT disturbance to ARMA$(2,1)$ to isolate trajectory effects.
Across RT configurations with varying temporal correlation, proposed method achieves the lowest contouring RMSE and MAE, indicating improved average geometric path fidelity under moving-reference perturbations.
However, the maximum contouring error is similar across controllers (about $50$--$53$\,mm), which suggests that the peak deviation is largely set by the disturbance transient itself rather than by steady tracking performance.

Fig.~\ref{fig:controlvalue} illustrates the adaptive regulation mechanism of the proposed framework. Following a disturbance, our MPCC formulation suppresses the virtual progress rate $v_s$ and drives the joint velocities toward their limits, thereby prioritizing geometric path recovery over temporal synchronization. This behavior reduces average contouring error, although the resulting transient lag can combine with the disturbance offset and yield slightly larger peak deviations in some cases.


\subsection{Physical Experiments}
\begin{figure}[t]
    \centering
    \includegraphics[width=0.7\linewidth]{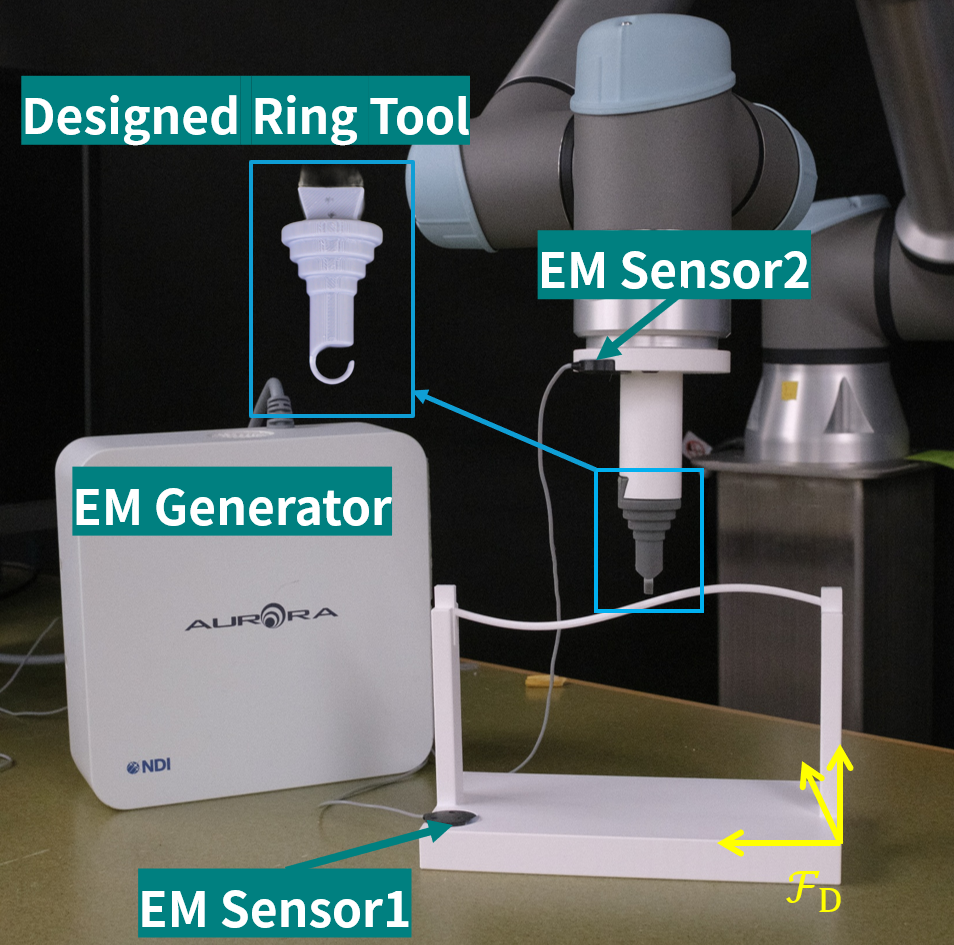}
    \caption{Physical validation setup. A $200$\,mm reference path is rigidly attached to a movable board defining the path frame $\mathcal{F}_D$.
    Two EM sensors measure the board pose ${}^{W}\mathbf{T}_{S_b}(t)$ and the tool pose ${}^{W}\mathbf{T}_{S_e}(t)$, enabling online estimation of
    ${}^{W}\mathbf{T}_{D}(t)$ and the end-effector pose ${}^{D}\mathbf{T}_{E}(t)$ for control and evaluation.}
    \label{fig:Physical_setup}
\end{figure}
\subsubsection{Experimental Setup}



\begin{figure}
    \centering
    \includegraphics[width=0.7\linewidth]{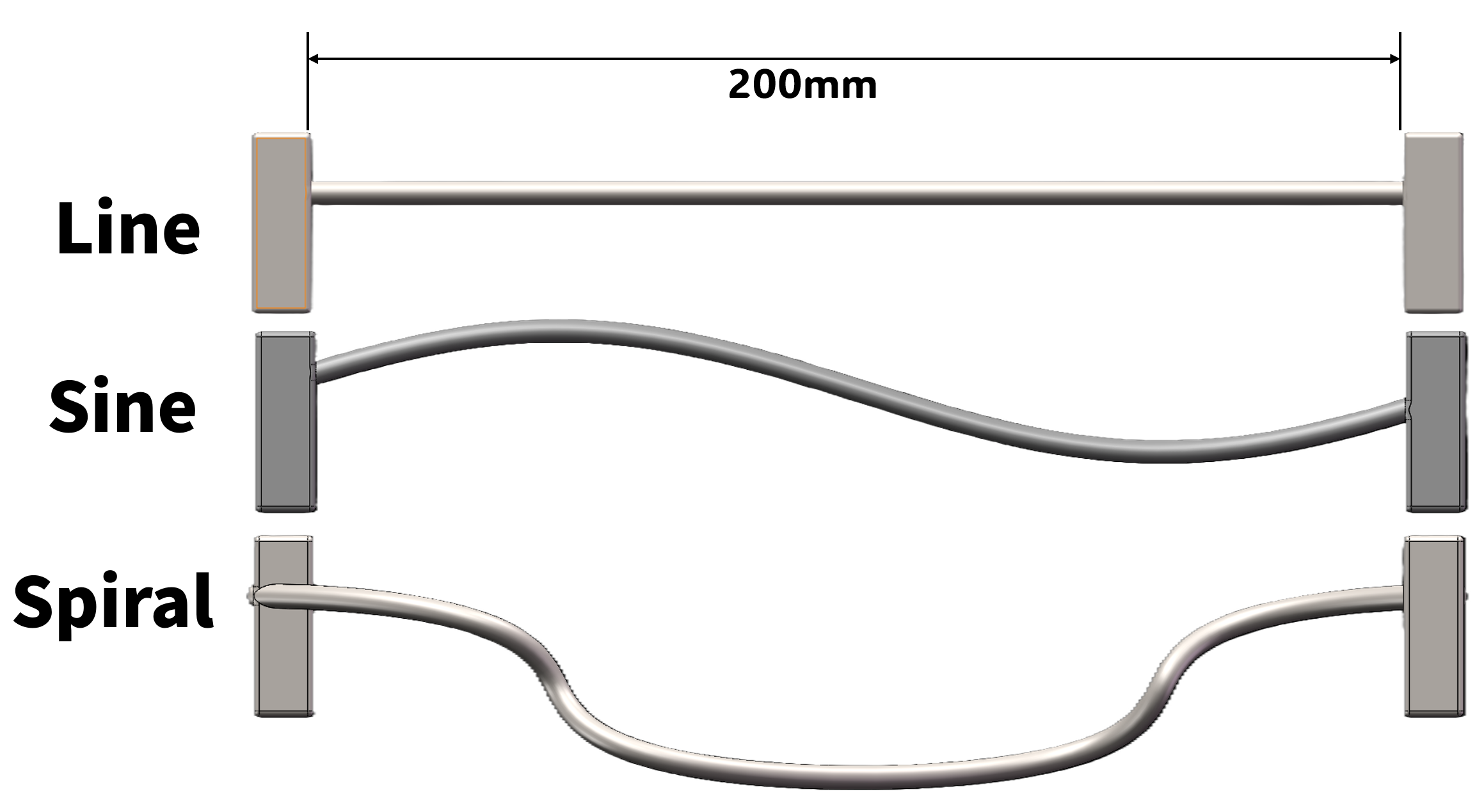}
    \caption{Trajectories used in physical experiment}
    \label{fig:Physical Trajectory}
\end{figure}

Physical validation is performed on a 6-DoF UR5e under control at $500$\,Hz. A planar board carries a $200$\,mm reference path (nominal width $4$\,mm) and defines the path frame $\mathcal{F}_D$, in which the geometric reference is fixed (Fig.~\ref{fig:Physical_setup}). Representative photographs of the straight, sine and spiral (same as in simulation) paths are shown in Fig.~\ref{fig:Physical Trajectory}.

An NDI Aurora electromagnetic (EM) tracking system at 40 Hz tracks the 6-DoF poses of a board-mounted sensor $S_b$ and a tool-mounted sensor $S_e$, with submillimetre positional and subdegree orientation accuracy \cite{li2023closed}.
Two constant transforms are identified offline: the hand--eye calibration ${}^{E}\mathbf{T}_{S_e}$ and the board-to-path registration ${}^{S_b}\mathbf{T}_{D}$.


During execution, the end-effector pose in the path frame is computed online from the EM measurements as
\begin{equation}
{}^{D}\mathbf{T}_{E}(t)
=
\left(
{}^{W}\mathbf{T}_{S_b}(t)\,{}^{S_b}\mathbf{T}_{D}
\right)^{-1}
{}^{W}\mathbf{T}_{S_e}(t)
\left({}^{E}\mathbf{T}_{S_e}\right)^{-1},
\end{equation}
where ${}^{W}\mathbf{T}_{S_b}(t)$ and ${}^{W}\mathbf{T}_{S_e}(t)$ are the EM measurements.
We also designed a ring tool (Fig.~\ref{fig:Physical_setup}) which is mounted at the end-effector as a physical tolerance gauge.

\subsubsection{Disturbance Protocol}
Disturbances are introduced by manually displacing the board during execution. A colored mark on each reference path indicates the predefined disturbance location; when the end-effector reaches this mark, the board is perturbed. This aligns disturbance onset with path progress rather than elapsed time.

The reference remains fixed in $\mathcal{F}_D$, while its pose in the robot base frame evolves according to the measured ${}^{W}\mathbf{T}_{D}(t)$.
For each run, disturbance magnitude is quantified post hoc from ${}^{W}\mathbf{T}_{D}(t)$ using the peak-to-peak translational and rotational deviations of the board motion.

\subsubsection{Evaluation Procedure}
Performance is evaluated in the path frame $\mathcal{F}_D$ using the estimated end-effector pose ${}^{D}\mathbf{T}_{E}(t)$.
Contouring error is computed along the executed trajectory relative to the geometric reference path.

\begin{figure}
    \centering
    \includegraphics[width=1.0\linewidth]{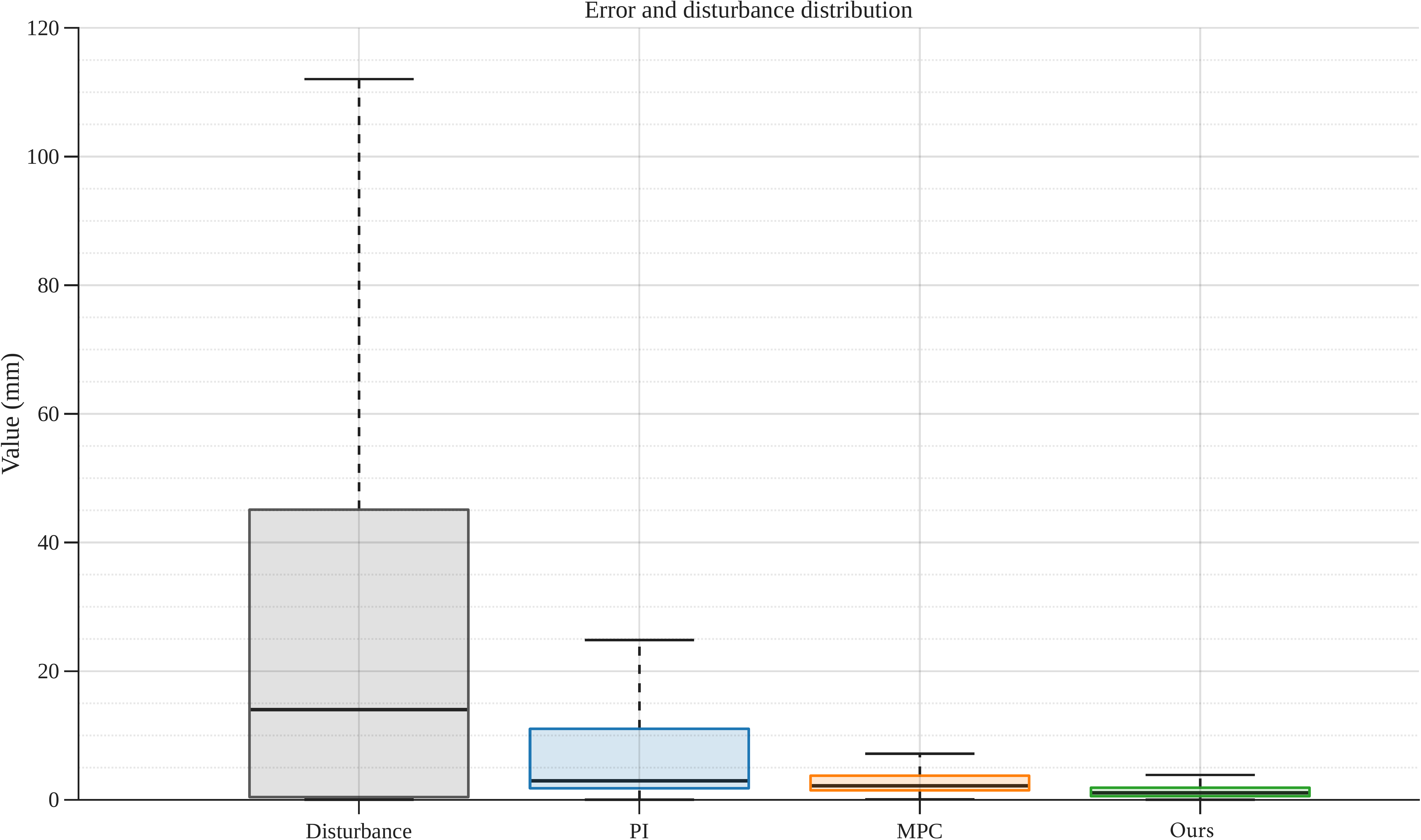}
    \caption{Boxplot of the applied moving-reference disturbance magnitude and the resulting contouring error distributions for PI, MPC, and Ours (evaluated in the path frame $\mathcal{F}_D$).}
    \label{fig:physical_boxplot}
\end{figure}
A run is deemed to violate the tolerance if the printed path contacts the ring tool, indicating a lateral deviation beyond the 16\,mm clearance. Three runs are conducted for each controller. The board motion is recorded by the EM tracking system and converted to the corresponding moving-reference disturbance, whose magnitude is summarized together with the contouring error distributions in Fig.~\ref{fig:physical_boxplot}.

\begin{figure*}[t]
    \centering
    \includegraphics[width=0.8\linewidth,trim={18cm 5.5cm 16.5cm 5.5cm},clip]{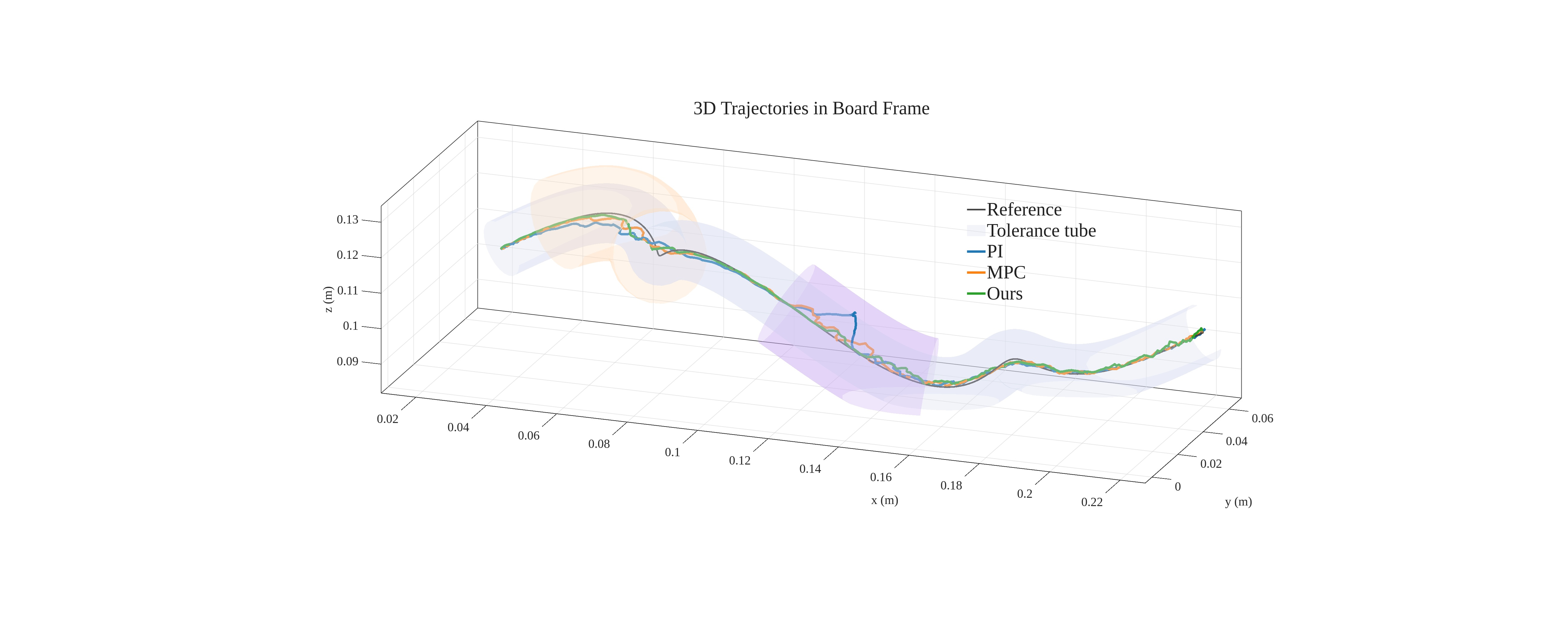}
    \caption{Representative reconstructed end-effector trajectories in the board-attached frame under manually applied moving-reference disturbances. Orange and purple shaded regions indicate the first and second disturbance intervals, respectively. The translucent tolerance tube is defined by the tool diameter. The proposed MPCC-based framework remains closer to the reference path than PI and time-indexed tracking MPC.}
    \label{fig:Physical_validation}
\end{figure*}

\subsubsection{Physical Validation Results}

Fig.~\ref{fig:Physical_validation} shows a representative reconstructed trajectory in the path frame, where deviations from the reference directly reflect geometric tracking error under board motion. After disturbance onset, Our framework remains the trajectory closer to the reference with less spatial spread, whereas PI shows larger deviations and tracking MPC exhibits intermediate performance.

To quantify performance beyond a single representative run, Fig.~\ref{fig:physical_boxplot} aggregates contouring error distributions across the three physical paths and ten runs per controller. The proposed MPCC-based framework exhibits a lower median contouring error and a smaller interquartile range than PI and tracking MPC across tested paths, indicating more consistent path relative accuracy under moving reference disturbances.
\section{Conclusion}
This paper proposed MR-MPCC, a hierarchical model predictive contouring control framework for full 6-DoF robotic manipulator path following under moving reference frames. The proposed method formulates manipulator tracking in a path-attached coordinate frame and optimizes the path progress online, allowing the controller to prioritize contouring accuracy when the reference frame undergoes translational and rotational motion. A constrained  resolved motion rate control based QP converts the commanded Cartesian twist into feasible joint velocities while enforcing joint-level and safety constraints.

MR-MPCC was evaluated in a unified ROS2--MuJoCo simulation and physical UR5e experimental pipeline. Moving-reference disturbances were applied directly to the path frame and measured using an electromagnetic tracking system. Under translation-only, rotation-only, and combined reference-frame disturbances, MR-MPCC consistently achieved lower contouring RMSE and MAE than PI control and time-indexed tracking MPC when evaluated relative to the moving path frame. The maximum contouring errors remained comparable across the controllers, indicating that the peak deviation was primarily governed by the transient reference-frame motion. During these transients, MR-MPCC reduced the optimized progress rate, improving average contouring accuracy at the expense of temporary lag accumulation.

The results demonstrate that MR-MPCC is effective for robotic manipulator path following when the desired path is defined relative to a moving reference frame. In particular, the online progress adaptation enables the controller to maintain more accurate path adherence without requiring strict synchronization with a time-indexed reference. Future work will incorporate sensing uncertainty, delayed and asynchronous measurements, and reference-frame state prediction into the estimation and control framework. The method will also be extended to contact-rich manipulation by introducing force-aware objectives and constraints.

\bibliography{reference}

@inproceedings{kurtz2021control,
  title={Control barrier functions for singularity avoidance in passivity-based manipulator control},
  author={Kurtz, Vince and Wensing, Patrick M and Lin, Hai},
  booktitle={2021 60th IEEE Conference on Decision and Control (CDC)},
  pages={6125--6130},
  year={2021},
  organization={IEEE}
}

@INPROCEEDINGS{Yoshikawa,
  author={Yoshikawa, T.},
  booktitle={Proceedings. 1985 IEEE International Conference on Robotics and Automation}, 
  title={Manipulability and redundancy control of robotic mechanisms}, 
  year={1985},
  volume={2},
  number={},
  pages={1004-1009},
  doi={10.1109/ROBOT.1985.1087283}}

@article{GARPINGER2014568,
  title   = {Performance and Robustness Trade-offs in PID Control},
  author  = {Garpinger, Olof and H{\"a}gglund, Tore and {\AA}str{\"o}m, Karl Johan},
  journal = {Journal of Process Control},
  volume  = {24},
  number  = {5},
  pages   = {568-577},
  year    = {2014},
  issn    = {0959-1524},
  doi     = {10.1016/j.jprocont.2014.02.020},
}

@ARTICLE{RMRC,
  author={Whitney, Daniel E.},
  journal={IEEE Transactions on Man-Machine Systems}, 
  title={Resolved Motion Rate Control of Manipulators and Human Prostheses}, 
  year={1969},
  volume={10},
  number={2},
  pages={47-53},
  doi={10.1109/TMMS.1969.299896}}

@inproceedings{li2023closed,
  title={A closed-form solution to electromagnetic sensor based intraoperative limb length measurement in total hip arthroplasty},
  author={Li, Tiancheng and Song, Yang and Walker, Peter and Pan, Kai and van de Graaf, Victor A and Zhao, Liang and Huang, Shoudong},
  booktitle={International Conference on Medical Image Computing and Computer-Assisted Intervention},
  pages={365--375},
  year={2023},
  organization={Springer}
}

@misc{pham2025robustadaptivefuzzysliding,
      title={Robust adaptive fuzzy sliding mode control for trajectory tracking for of cylindrical manipulator}, 
      author={Van Cuong Pham and Minh Hai Tran and Phuc Anh Nguyen and Ngoc Son Vu and Nga Nguyen Thi},
      year={2025},
      eprint={2508.05584},
      archivePrefix={arXiv},
      primaryClass={cs.RO}, 
}

@article{FIGUEREDO2021109817,
title = {Robust {$H_\infty$} kinematic control of manipulator robots using dual quaternion algebra},
journal = {Automatica},
volume = {132},
pages = {109817},
year = {2021},
issn = {0005-1098},
doi = {https://doi.org/10.1016/j.automatica.2021.109817},
author = {Luis Felipe da Cruz Figueredo and Bruno Vilhena Adorno and João Yoshiyuki Ishihara}
}

@article{kapitanyuk2017guiding,
  title={A guiding vector-field algorithm for path-following control of nonholonomic mobile robots},
  author={Kapitanyuk, Yuri A and Proskurnikov, Anton V and Cao, Ming},
  journal={IEEE Transactions on Control Systems Technology},
  volume={26},
  number={4},
  pages={1372--1385},
  year={2017},
  publisher={IEEE}
}

@article{ji2019industrial,
  title={Industrial robotic machining: a review},
  author={Ji, Wei and Wang, Lihui},
  journal={The International Journal of Advanced Manufacturing Technology},
  volume={103},
  number={1},
  pages={1239--1255},
  year={2019},
  publisher={Springer}
}

@incollection{faulwasser2009nonlinear,
  title={Nonlinear model predictive path-following control},
  author={Faulwasser, Timm and Findeisen, Rolf},
  booktitle={Nonlinear Model Predictive Control: Towards New Challenging Applications},
  pages={335--343},
  year={2009},
  publisher={Springer}
}

@INPROCEEDINGS{9873215,
  author={Zhou, Zhiwei and Gao, Jian and Zhang, Lanyu},
  booktitle={2022 23rd International Conference on Electronic Packaging Technology (ICEPT)}, 
  title={Research on Contour Tracking Control Method of Three-axis Motion Platform}, 
  year={2022},
  volume={},
  number={},
  doi={10.1109/ICEPT56209.2022.9873215}}

@article{liFiveaxisContourError2023,
  title   = {Five-Axis Contour Error Control Based on Spatial Iterative Learning},
  author  = {Li, Jiangang and You, Zhiyang and Li, Yanan and Miao, Enming and Yue, Ruijie},
  journal = {IEEE Transactions on Automation Science and Engineering},
  volume  = {20},
  pages   = {112--123},
  year    = {2023},
  doi     = {10.1109/TASE.2022.3142565},
}

@article{Yoon2025ReactiveMP,
  title={Reactive Model Predictive Contouring Control for Robot Manipulators},
  author={Junheon Yoon and Woo-Jeong Baek and Jaeheung Park},
  journal={2025 IEEE/RSJ International Conference on Intelligent Robots and Systems (IROS)},
  year={2025},
  pages={365-372},
}

@article{romeroModelPredictiveContouring2022,
  title   = {Model Predictive Contouring Control for Time-Optimal Quadrotor Flight},
  author  = {Romero, Angel and Sun, Sihao and Foehn, Philipp and Scaramuzza, Davide},
  journal = {IEEE Transactions on Robotics},
  volume  = {38},
  number  = {6},
  pages   = {3340--3356},
  year    = {2022},
  doi     = {10.1109/TRO.2022.3173711},
}

@inproceedings{lamModelPredictiveContouring2010,
  title     = {Model Predictive Contouring Control},
  author    = {Lam, Denise and Manzie, Chris and Good, Malcolm},
  booktitle = {Proc. IEEE Conf. on Decision and Control (CDC)},
  pages     = {6137--6142},
  year      = {2010},
  doi       = {10.1109/CDC.2010.5717042}
}

@article{moore1992iterative,
  title={Iterative learning control for trajectory tracking of robot manipulators},
  author={Moore, Kevin L.},
  journal={IEEE Transactions on Industrial Electronics},
  volume={39},
  number={1},
  pages={1--9},
  year={1992},
  doi={10.1109/41.125466}
}

@article{bristow2006survey,
  title={A survey of iterative learning control},
  author={Bristow, Douglas A. and Tharayil, Marina and Alleyne, Andrew G.},
  journal={IEEE Control Systems Magazine},
  volume={26},
  number={3},
  pages={96--114},
  year={2006},
  doi={10.1109/MCS.2006.1636313}
}

@article{linigerOptimizationbasedAutonomousRacing2015a,
  title   = {Optimization-Based Autonomous Racing of 1:43 Scale {RC} Cars},
  author  = {Liniger, Alexander and Domahidi, Axel and Morari, Manfred},
  journal = {Optimal Control Applications and Methods},
  volume  = {36},
  number  = {5},
  pages   = {628--647},
  year    = {2015},
  doi     = {10.1002/oca.2123},
}

@software{woolfrey_simple_qp_2023,
     author  = {Woolfrey, Jon},
     month   = aug,
     title   = {{S}imple {QP} {S}olver},
     url     = {https://github.com/Woolfrey/software_simple_qp},
     version = {1.0.0},
     year    = {2023}
}

@misc{jiCMPCCCorridorbasedModel2021,
  title        = {{CMPCC}: Corridor-based Model Predictive Contouring Control for Aggressive Drone Flight},
  author       = {Ji, Jialin and Zhou, Xin and Xu, Chao and Gao, Fei},
  year         = {2021},
  eprint       = {2007.03271},
  archivePrefix= {arXiv},
  primaryClass = {cs},
  doi          = {10.48550/arXiv.2007.03271},
}

@ARTICLE{astudillo2022tunnelfollowing,
  author={Astudillo, Alejandro and Gillis, Joris and Diehl, Moritz and Decré, Wilm and Pipeleers, Goele and Swevers, Jan},
  journal={IEEE Robotics and Automation Letters}, 
  title={Position and Orientation Tunnel-Following NMPC of Robot Manipulators Based on Symbolic Linearization in Sequential Convex Quadratic Programming}, 
  year={2022},
  volume={7},
  number={2},
  pages={2867-2874},
  doi={10.1109/LRA.2022.3142396}}

@article{romero2022time,
  title={Time-optimal online replanning for agile quadrotor flight},
  author={Romero, Angel and Penicka, Robert and Scaramuzza, Davide},
  journal={IEEE Robotics and Automation Letters},
  volume={7},
  number={3},
  pages={7730--7737},
  year={2022},
  publisher={IEEE}
}

@book{craigIntroductionRoboticsMechanics2018,
  title     = {Introduction to Robotics: Mechanics and Control},
  author    = {Craig, John J.},
  year      = {2018},
  edition   = {4th},
  publisher = {Pearson}
}

@article{yao2023disturbance_tmmpc,
  title   = {Inverse-Dynamics- and Disturbance-Observer-Based Tube Model Predictive Tracking Control of Uncertain Robotic Manipulators},
  author  = {Yao, Xiuming and Xu, Hongze and Yang, Yu},
  journal = {Journal of the Franklin Institute},
  volume  = {360},
  number  = {10},
  pages   = {6906--6927},
  year    = {2023},
  doi     = {10.1016/j.jfranklin.2023.04.005}
}

@article{kang2023sm_tube_mpc,
  title   = {Sliding Mode-Based Adaptive Tube Model Predictive Control for Robotic Manipulators with Model Uncertainty and State Constraints},
  author  = {Kang, Erlong and Liu, Yang and Qiao, Hong},
  journal = {Control Theory and Technology},
  volume  = {21},
  pages   = {334--351},
  year    = {2023}
}

@book{camachoMPC2007,
  title     = {Model Predictive Control},
  author    = {Camacho, Eduardo F. and Bordons, Carlos},
  edition   = {2nd},
  publisher = {Springer},
  year      = {2007},
  doi       = {10.1007/978-0-85729-398-5}
}

@article{hewingLearningBasedMPC2020,
  title   = {Learning-Based Model Predictive Control: Toward Safe Learning in Control},
  author  = {Hewing, Lukas and Wabersich, Kim P. and Menner, Marcel and Zeilinger, Melanie N.},
  journal = {Annual Review of Control, Robotics, and Autonomous Systems},
  volume  = {3},
  pages   = {269--296},
  year    = {2020},
}

@article{woolfreyPredictiveEndEffectorControl2021,
  title   = {Predictive End-Effector Control of Manipulators on Moving Platforms Under Disturbance},
  author  = {Woolfrey, Jon and Lu, Wenjie and Liu, Dikai},
  journal = {IEEE Transactions on Robotics},
  volume  = {37},
  number  = {6},
  pages   = {2210--2217},
  year    = {2021},
  doi     = {10.1109/TRO.2021.3072544}
}

@book{siciliano2009handbook,
  title     = {Springer Handbook of Robotics},
  editor    = {Siciliano, Bruno and Khatib, Oussama},
  publisher = {Springer},
  year      = {2009},
  doi       = {10.1007/978-3-540-30301-5}
}

\end{document}